\documentclass[
twocolumn,
numbers, % Enable numbered citations
sort&compress, % Sort and compress citation ranges
]{ceurart}

\usepackage{listings}

\usepackage{hyperref}
\usepackage{url}
\usepackage{enumitem}
\usepackage{verbatim}
\usepackage{multicol}
\usepackage{lipsum}

\begin{document}

\copyrightyear{2024}
\copyrightclause{Copyright for this paper by its authors.
  Use permitted under Creative Commons License Attribution 4.0
  International (CC BY 4.0).}

\conference{Educational Data Mining 2024 Workshop: Leveraging Large Language Models for Next Generation Educational Technologies, July 14, 2024, Atlanta, Georgia, USA}

\title{SPL: A Socratic Playground for Learning Powered by Large Language Model}

\tnotemark[1]

\author[1,2]{Liang Zhang}[%
orcid=0009-0002-0017-2569,
email=lzhang13@memphis.edu
]
\address[1]{Institute for Intelligent Systems, University of Memphis, Memphis, TN 38152, USA}
\address[2]{Department of Electrical and Computer Engineering, University of Memphis, Memphis, TN 38152, USA}

\author[3]{Jionghao Lin}[%
orcid=0000-0003-3320-3907,
email=jionghao@cmu.edu]
\address[3]{School of Computer Science, Carnegie Mellon University, Pittsburgh, PA, 15213, USA}

\author[4]{Ziyi Kuang}[%
email=ziyikuangzyk@gmail.com
]
\cormark[1]
\address[4]{School of Psychology, Shaanxi Normal University, Xi’an, 710062, PR China}

\author[5]{Sheng Xu}[%
email=willxusheng@gmail.com
]
\address[5]{School of Psychology, Central China Normal University, Wuhan, 430079, PR China}

\author[6]{Xiangen Hu}[%
email=xiangen.hu@polyu.edu.hk
]
\address[6]{Department of Applied Social Sciences, Hong Kong Polytechnic University, Hong Kong, PR China}

\cortext[1]{Corresponding author.}
% \fntext[1]{These authors contributed equally.}

\begin{abstract}
  Dialogue-based Intelligent Tutoring Systems (ITSs) have significantly advanced adaptive and personalized learning by automating sophisticated human tutoring strategies within interactive dialogues. However, replicating the nuanced patterns of expert human communication remains a challenge in Natural Language Processing (NLP). Recent advancements in NLP, particularly Large Language Models (LLMs) such as OpenAI's GPT-4, offer promising solutions by providing human-like and context-aware responses based on extensive pre-trained knowledge. Motivated by the effectiveness of LLMs in various educational tasks (e.g., content creation and summarization, problem-solving, and automated feedback provision), our study introduces the Socratic Playground for Learning (SPL), a dialogue-based ITS powered by the GPT-4 model, which employs the Socratic teaching method to foster critical thinking among learners. Through extensive prompt engineering, SPL can generate specific learning scenarios and facilitates efficient multi-turn tutoring dialogues. The SPL system aims to enhance personalized and adaptive learning experiences tailored to individual needs, specifically focusing on improving critical thinking skills. Our pilot experimental results from essay writing tasks demonstrate SPL has the potential to improve tutoring interactions and further enhance dialogue-based ITS functionalities. Our study, exemplified by SPL, demonstrates how LLMs enhance dialogue-based ITSs and expand the accessibility and efficacy of educational technologies. 
\end{abstract}


\maketitle

\section{Introduction}

Dialogue-based Intelligent Tutoring Systems (ITSs) leverage artificial intelligence to simulate human-like tutoring through interactive dialogues \cite{nye2014autotutor,paladines2020systematic}. These systems aim to provide personalized and adaptive learning experiences by engaging learners in conversation, such as asking questions and providing feedback, and guiding them towards the expected learning goals. Over the past three decades, dialogue-based ITSs have demonstrated effectiveness in supporting learning, particularly in STEM subjects \cite{graesser2023intelligent} as well as in reading and language learning \cite{graesser2001intelligent,nye2014autotutor,paladines2020systematic}. However, dialogue-based ITSs can still be improved by incorporating more human-like guidance (e.g., effective tutoring strategies and polite language \cite{paladines2020systematic, lin2022good, lin2022exploring}), which underscores the importance of fully replicating the nuanced patterns of expert human tutoring communication within ITSs. In this context, advancements in large language models (LLMs) offer promising solutions, such as in-context learning and detailed feedback, for enhancing the quality of instruction \cite{xu2024context, lin2024can,stamper2024enhancing}.

LLMs, such as OpenAI's GPT-4 \cite{achiam2023gpt}, are pre-trained on extensive datasets and can generate human-like dialogue when properly prompted. These models leverage their vast knowledge base to exhibit human-like reasoning and deliver insightful responses in natural language \cite{brown2020language}. A critical technique for maximizing the capabilities of LLMs is prompt engineering, which includes methods like chain-of-thought (CoT) prompting \cite{wei2022chain} and few-shot prompting \cite{brown2020language}. These methods enhance the models' ability to replicate human interaction and provide more adaptive text generation. Previous research has highlighted the promise of LLMs in improving various educational tasks, including providing better feedback \cite{dai2023can,stamper2024enhancing}, enhancing learning guidance and interaction strategies \cite{kumar2023impact}, understanding student behaviors \cite{park2024empowering}, and stimulating tutoring dialogues through answer evaluation and content generation \cite{nye2023generative}.

\begin{figure*}[ht!]
\centering
\includegraphics[height=3in, width=6.1in]{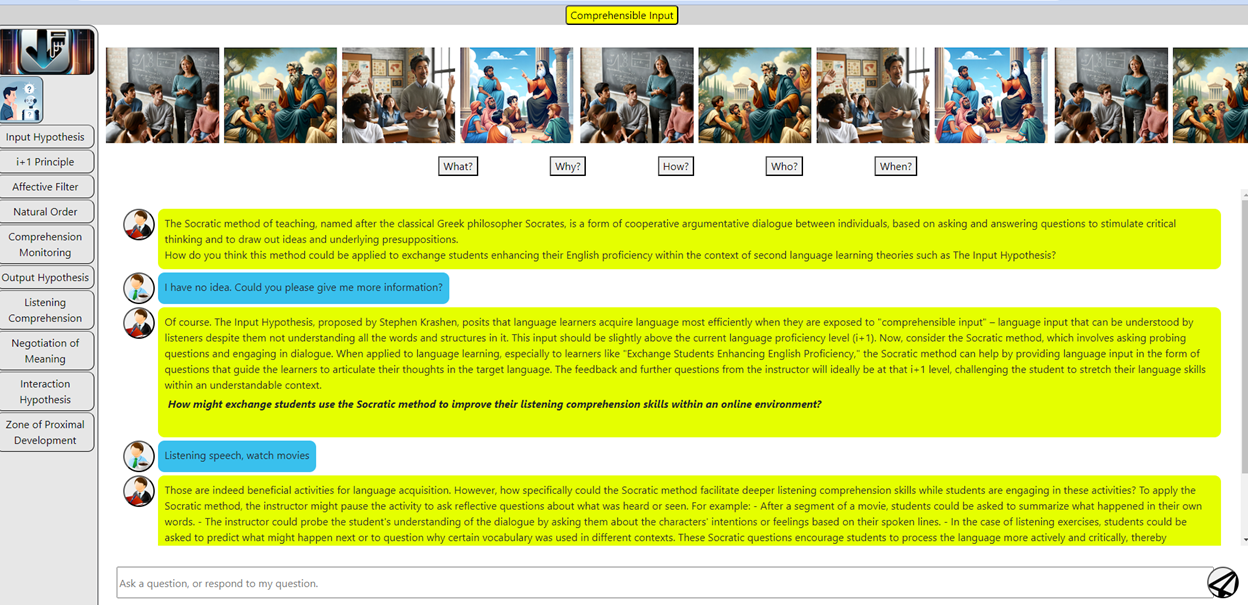}
\caption{An Example Interactive Dialogue Interface of SPL System.}
\label{interface}
\end{figure*}

Inspired by the potential of LLMs in education, our study introduces a dialogue-based Intelligent Tutoring System (ITS) named the Socratic Playground for Learning  (SPL)\footnotemark[1]\footnotetext[1]{SPL Platform: \href{https://polyu.skoonline.org/}{https://polyu.skoonline.org/}}, which simulates the Socratic teaching method in specific learning scenarios. SPL guides learners to solve questions by fostering self-reflection, critical thinking, and the development of independent thinking skills through interactive dialogue \cite{yang2023socratic,kong2023platolm}. Leveraging the capability of GPT models with advanced prompt engineering, SPL aims to deliver adaptive and flexible learning experience that can adjust to various educational contexts and learner profiles. Figure \ref{interface} illustrates an example of user interface for SPL dialogue, designed to enhance English proficiency for learners by applying second language learning principles. The interface features a menu with multiple selectable learning principles (e.g., Zone of Proximal Development) in the left-side column and five types of wh-questions (What?, Why?, How?, Who?, When?) at the top. This design enables learners to engage in interactive dialogues that promote critical thinking and language acquisition. The initial dialogue developed by the SPL system begins with the question ``\textit{How do you think this method could be applied to exchange students enhancing their English proficiency within the context of second language learning theories such as The Input Hypothesis?}''. This is followed by a multi-turn dialogue with additional prompt wh-questions to further stimulate the learner's critical thinking. 

Our study introduces the Socratic Playground for Learning (SPL) system, which employs GPT-4-based prompt strategies to create personalized learning scenarios grounded in the Socratic teaching method, enhancing dialogue-driven educational interactions. SPL demonstrates a significant enhancement over traditional dialogue-based ITSs by automating lesson design for specific learning scenarios and utilizing sophisticated NLP capabilities for multi-turn dialogue tutoring, thereby reducing reliance on human effort and predefined rules. Our preliminary evaluation of the SPL system's capabilities was conducted using essay writing tasks with college students. The results demonstrate the positive impact of the system's effective use of LLM in facilitating learning through the Socratic teaching method, promoting both critical thinking and deeper comprehension. Additionally, SPL provides adaptive and flexible learning experiences, increasing scalability and enabling the system to adjust to various educational contexts and learner profiles, thus broadening its potential for widespread adoptions in AI-based education. 

\section{Related Work}
\subsection{Dialogue-based Intelligent Tutoring Systems}

Dialogue-based ITSs have proven to be effective in fostering cognitive engagement and improving learning outcomes by utilizing conversational interactions modeled on the best practices of human tutors \cite{feng2021systematic,d2023intelligent}. Since the development of the early SCHOLAR tutor by Carbonell in 1970, which offered Socratic tutoring through natural language text input and output \cite{carbonell1970ai}, dialogue-based ITSs have employed mixed-initiative dialogues, semantic networks, question-answering, and tailored feedback to enhance learning. The SCHOLAR system encouraged learners to both ask and answer questions, providing feedback based on their responses to guide them toward the correct answers. Despite these advances, fully replicating all the capabilities of a human tutor remains a distant goal due to persistent challenges in natural language processing techniques \cite{paladines2020systematic}.

The development of AutoTutor marked a significant advancement in dialogue-based ITSs by incorporating tutoring strategies derived from human tutoring protocols \cite{graesser1999autotutor}. AutoTutor poses questions and problems from a curriculum script, understands learner inputs entered via keyboard, generates tutoring strategies in response (such as brief feedback, prompts, elaborations, corrections, and hints), and presents these strategies through a talking head \cite{graesser1999autotutor,graesser2001intelligent}. The dialogue structure in AutoTutor is guided by the expectation-and-misconception-tailored (EMT) dialogue rule, a pedagogical method for scaffolding student answers. Later on, many dialogue-based ITSs have been developed for diverse subjects. For example, Why2-Atlas is a natural language-based ITS for qualitative physics that uses deep syntactic analysis and abductive theorem proving to identify and address misconceptions in students' explanatory essays through dialogue-based feedback \cite{vanlehn2002architecture}. The Geometry Explanation Tutor engages students in dialogue-based self-explanation to improve their understanding and articulation of geometry rules \cite{aleven2004evaluating}. DeepTutor is a conversational ITS that aligns assessment, learning progressions, and instructional tasks to guide students through conceptual physics problems with personalized instruction and feedback \cite{rus2014deeptutor,rus2015deeptutor}.

\subsection{LLMs for Enhancing ITSs}

Large language models, such as ChatGPT, have brought opportunities to the ITS community in areas such as lesson design, feedback generation, and assessment of learner knowledge mastery. Ahmed \cite{ahmed2023chatgpt} explored the potential of ChatGPT for conversation design and assessment in a course, facilitating the Generalized Intelligent Framework for Tutoring (GIFT) and reducing the effort required to design EMT conversation scripts. The conversational tutoring system Ruffle \& Riley, developed by Schmucker et al., automatically generates tutoring scripts from lesson texts using GPT-4 to  accelerate content authoring and employs the EMT-based rules to facilitate free-form conversational tutoring \cite{schmucker2023ruffle, schmucker2024ruffle}. Abu-Rasheed et al. \cite{abu2024supporting} proposed an LLM-based chatbot that engages students in conversation, similar to a discussion with a peer or mentor, augmented with knowledge graphs and human mentorship, enhancing conversational explainability (e.g., clarifying the reasons behind specific
content suggestions) and mentoring in educational recommendations. Dan et al. \cite{dan2023educhat} developed EduChat, a large-scale language model-based chatbot system for intelligent education that provides personalized, comprehensive and timely support for teachers, students, and parents by integrating retrieval-augmented question-answering, essay assessment, Socratic teaching, and emotional support to facilitate personalized and compassionate learning, leveraging pre-trained knowledge from educational and psychological domains. Nye et al. \cite{nye2023generative} highlighted opportunities for enhancing educational experiences through content generation with LLMs, while also addressing concerns around inaccuracies and equitable access.

Recent advancements in large language models (LLMs) have driven further innovation in education. Dai et al. \cite{dai2023can, daiassessing} demonstrated that GPT models could automate students' performance assessment and feedback generation in a manner more readable than that of human tutors. Lin et al. \cite{lin2024i} developed a GPT-4-powered feedback system that  that provides explanatory feedback by identifying trainees' responses as desired or undesired and automatically generating template-based feedback, with the GPT-4 model rephrasing incorrect responses to ensure clarity and understanding. Zhang et al. \cite{zhang2024predicting} explored the potential of LLMs in predicting learning performance, finding that they outperform traditional knowledge tracing methods in predictive accuracy in the context of adult literacy. 

\section{Socratic Playground for Learning (SPL)}

The SPL is an LLM-powered, dialogue-based ITS designed to facilitate in-context learning through the Socratic teaching method \cite{SPL_FAQ}. It uses standard prompt strategies for lesson creation and Socratic dialogue to stimulate critical thinking and uncover underlying ideas and assumptions. The SPL offers personalized, adaptive, and flexible learning experiences that promote self-reflection and the development of critical thinking skills in learners. 

\subsection{System Architecture}

\begin{figure*}[h!]
\centering
\includegraphics[height=2.0in, width=6.1in]{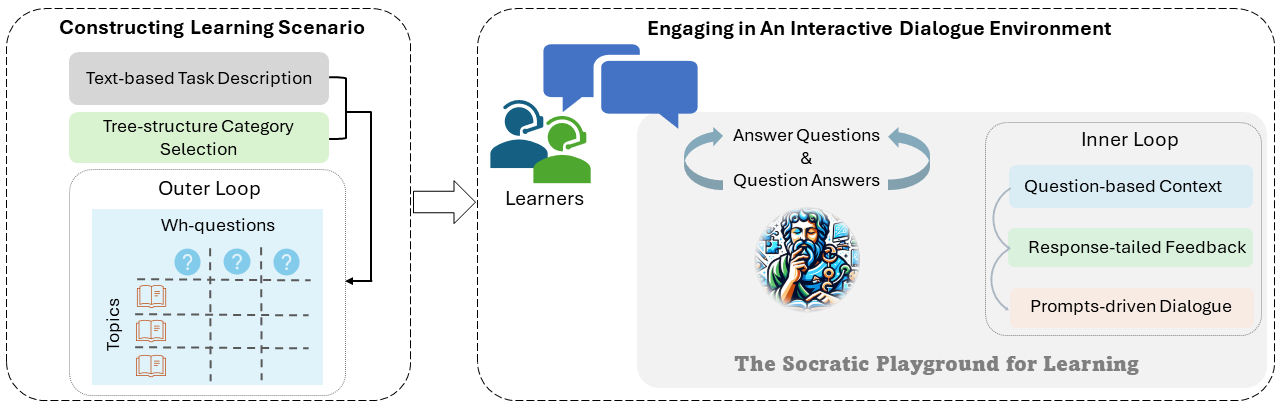}
\caption{The Socratic Playground for Learning  System Architecture.}
\label{architecture}
\end{figure*}

As illustrated in Figure \ref{architecture}, the SPL system architecture supports usage in two main stages: (1) constructing learning scenarios and (2) engaging in an interactive dialogue environment. 

\textbf{Constructing Learning Scenarios.} This system allows both educators and learners to easily and automatically construct personalized learning scenarios. Users can create scenarios through text-based task descriptions or by selecting options from a tree-structured format, which includes hierarchical categories in up-down relations such as domain, subdomain, objective, context, concepts, target learners, environments, and tutoring pedagogies. For example, a user might describe their learning request as: ``\textit{I am John, struggling with time management affecting grades. Any tips on effective time management would be welcome.}'' Alternatively, they can as select from a tree structure: choosing ``Psychology'' as the domain, ``Educational Psychology'' as the subdomain, and setting the goal as ``To understand the impact of motivation on student learning.''. Based on this input, the system, powered by LLM (GPT-4), generates a list of learning contexts, such as ``Explore the role of extrinsic rewards in student motivation.''. The system then generates a corresponding list of concepts, such as the ``Behavior Reinforcement", to be chosen as main focus. ``College Students'' is chosen from the target learner list, identifying the primary audience, and "Online Discussions" is selected from the learning environments list, indicating the mode of interaction. Finally, users can select from the list of pedagogical strategies, such as the ``Socratic Method'', though other methods like BLOOM (tutoring concepts/skills at all 6 levels of Bloom), TIMSS (Trends in International Mathematics and Science Study that tutoring based on different cognitive domains), Game-based learning (e.g., Who wants to be Millionaire), and Teachable Agents (OpenAI needs your help to understand the concepts) are also available. The up-down tree structure category selection process provides the necessary information, knowledge, and background to facilitate the automatic construction of learning scenarios. The established scenarios displayed on the SPL user interface showcase a matrix format of knowledge components or topics derived from input information and wh-questions (e.g., ``what?'', ``why?'', ``how?'', ``who?'', ``when?'') \cite{koshik2003wh,fox2010responses}. By selecting and integrating these two dimensions, use can refine creation process, initiate the defined learning scenario, and start with specific questions to provoke dialogue, targeting their preferred knowledge areas (as illustrated in the outer loop in Figure \ref{architecture}). For example the questions like ``\textit{What} effect do you think earning badges for your contributions might have on your motivation to participate in the online discussions?'' and ``\textit{How} do you think earning badges for your contributions might impact your motivation to participate in the online discussions?". 

\textbf{Engaging in An Interactive Dialogue Environment.} Within the dialogue interface, users engage in interactive dialogues driven by the Socratic method (as illustrated in the inner loop of the SPL architecture in Figure \ref{architecture}). The process includes: 1) \textit{Question-based Context:} Initially, the SPL system kicks off the dialogue by presenting an example scenario or context related to the specific task or knowledge and posing a wh-question about that context. 2) \textit{Learner Response-tailed Feedback:} After the user responds, the system captures the learners' responses or historical records, assesses their understanding, and provides timely feedback concluding with an another prompt question to encourage deeper thinking. 3) \textit{Iterative Prompts-driven Dialogue:} The system persistently guides users through iterative prompts, deepening their thinking, correcting errors, and leading to correct solutions, providing a dynamic and interactive experience as a key pedagogical strategy within the dialogue-based environment. For example, considering the established learning scenario: ``\textit{Imagine a student named Taylor who has set a goal to improve their grades this semester. Taylor is exploring different motivational strategies to stay on track and achieve this goal. Chart your path to success by mastering the art of motivation. Let's embark on this journey together!}''. The question ``\textit{What motivational strategies do you think Taylor could use to achieve their goal?}'' kicks off the multiple-turn dialogue. If the user responds with ``\textit{I believe it requires hard work}'', the system might reply, ``\textit{Absolutely, hard work is essential. But let's dive deeper into specific strategies that can help Taylor stay motivated.}''. Here, the word ``\textit{Absolutely}'' asserts agreement by assessing the user's response and provides positive feedback. This is further followed by the prompt question, ``\textit{What types of positive reinforcement could Taylor use to maintain their motivation and improve their grades?}''. If the user's response is, ``\textit{I think some verbal praise and goal setting.}'', the system would follow up with, ``\textit{Great start! Verbal praise and goal setting can be powerful motivators.}'', and then prompt again, ``\textit{How do you think verbal praise can impact Taylor's motivation and academic performance?}''. This approach both validates the user's response again and encourages deeper thinking and elaboration on ``\textit{how}'' aspect. The iterative loop continues with diverse wh-questions, fostering critical thinking and deeper engagement for learners. 

\subsection{System Prompt Engineering}
The entire process, including the scenario construction and multiple-turn interactive dialogue, is driven by GPT-4 based prompt engineering, which supports the Socratic teaching method for learning in the SPL. Several important nodes are described below:

\textbf{Standard Prompt for Lesson Creation.} This approach structures the creation of educational scenarios by starting with broad knowledge areas and refining them into specific sub-components. Leveraging GPT-4's reasoning, knowledge, prediction, and generative abilities, it transitions from general concepts to detailed elements essential for generating specific scenarios. This method effectively navigates complex information, facilitating the construction of learning scenarios, including role definitions, task clarifications, context setting, content specification, question generation, instructional resource preparation, pedagogical approach selection, and detailed scenario development. See the Table \ref{tab:prompt_template} as an example structure of a standard prompt template for lesson creation, demonstrating how broad concepts are refined into specific, actionable components. This systematic approach ensures clarity and precision in generating learning scenarios, fostering an effective and engaging learning environment. The prompt defines some variables, which are detailed below:

\begin{itemize}[leftmargin=*,itemsep=0pt, topsep=0pt,partopsep=0pt, parsep=0pt]
    \item \texttt{\%[theLang]\%} refers to the language (e.g., English, Chinese Mandarin, etc.) that will be displayed in the SPL learning scenario. This ensures that the content is accessible to learners in their preferred language. 
    \item \texttt{\%[theKC]\%} refers to the knowledge components required to constitute the knowledge space for the domain-specific scenario. These components are essential elements or concepts that form the foundation of the subject matter. 
    \item \texttt{\%[theNumber]\%} refers to the number of concepts needed for the creation of the learning scenario. This helps in defining the scope and depth of the learning material. 
    \item \texttt{\%[theDomain]\%} refers to the domains (e.g., computer science, business, psychology, etc.) used in creating the specific learning scenario. This specifies the academic or professional field to which the learning scenario belongs.  
    \item \texttt{\%[theTarget]\%} refers to the target learner group (e.g., college students, graduate students, online learners, etc.). This identifies the primary audience for the learning scenario, ensuring that the content is tailored to their needs and level of understanding.
    \item \texttt{\%[theAvatar]\%} refers to the avatar displayed in the user interface of the SPL dialogue. This personalized character can enhance engagement and provide a more interactive learning experience. 
    \item \texttt{\%[theTutorName]\%} refers to the name that users prefer for the virtual tutor, adding a personalized touch to the tutoring experience.
    \item \texttt{\%[theContext]\%} refers to the context by topics for learning scenarios. This specifies the thematic areas or situations that the learning material will address. 
    \item \texttt{\%[theEnvironment]\%} refers to the learning environment (e.g., online learning) used for learning engagement. This defines the setting in which the learning activities will take place, influencing the methods and tools used for instruction. 
        \item \texttt{\%[theUserName]\%} refers to the user name for designing the SPL learning scenario. This personalizes the experience and can be used for tracking progress and providing feedback. 
        \item \texttt{\%[theType]\%} refers to the style of pedagogical strategies for the learning scenario, e.g., Socratic method. This defines the instructional approach used to facilitate learning and ensure the material is effectively delivered. 
    \item \texttt{\%[theObjective]\%} refers to the goal set for the learning, e.g., understanding the principles of working memory and understanding problem-solving strategies, in a specific learning subject. %Clear objectives help guide the learning process and measure success. 
\end{itemize}

% \noindent\hspace*{2cm}
\begin{table}[ht!]
\caption{Standard Prompt Template for Lesson Creation.}
\centering
\scriptsize
\renewcommand{\arraystretch}{1.5}
\begin{tabular}{|p{7cm}|}
\hline

Your answers, both for now and for future interactions, will be presented in \texttt{\%[theLang]\%}. \\

You are producing some basic concepts, called knowledge components relevant to \texttt{\%[theKC]\%}, in \texttt{\%[theDomain]\%} for a group of \texttt{\%[theTarget]\%}. \\

Please give me \texttt{\%[theNumber]\%} concepts relevant to \texttt{\%[theDomain]\%}. output each separately, in pure json, following this format:
\begin{verbatim}
{
  "theAvatar":"%[theAvatar]%",
  "theLang":"%[theLang]%",
  "theKC":%[the_concept]%,
  "theType":"%[theType]%",
  "theTarget":"%[theTarget]%",
  "theTutorName":"%[theTutorName]%",
  "theContext":"%[theContext]%",
  "theEnvironment":"%[theEnvironment]%",
  "theUserName":"%[theUserName]%",
  "theStyle":"%[theType]%",
  "theObjective":"%[theObjective]%"
}
\end{verbatim}

Making sure each of the entry in its own, pure, json. \\

Do not put all in one array. one json for each of \texttt{\%[theNumber]\%} concepts. \\

And the last but not least, making sure the value of the json objects are in \texttt{\%[theLang]\%}. in English only if you are not sure. \\

Make \texttt{theKC} short (less than 3 words if the language is English). 

 \\ \hline
\end{tabular}
\label{tab:prompt_template}
\end{table}

\textbf{Standard Prompt for Interactive Socratic Dialogue.} The SPL employs a carefully designed prompt architecture powered by GPT-4, integrating the Socratic method to foster interactive and engaging tutoring. The design rules are implemented through prompt templates, which are dynamically updated based on the dialogue interactions. For more details, please refer to Table \ref{tab:SocraticDialogue}. The table outlines various prompt types involved the interaction process, their descriptions, and example wh-questions, showcasing how the system guides learners through context, feedback, and iterative questioning. This structured approach ensures a personalized and adaptive learning experience, encouraging critical thinking and reflection. 

\begin{table*}[ht!]
\centering

\caption{Standard Prompt for Interactive Socratic Dialogue.}
\scriptsize
\renewcommand{\arraystretch}{1.4}
\begin{tabular}{p{3.0cm}|p{6cm}|p{5.2cm}}
\hline
\textbf{Prompt Type} & \textbf{Description} & \textbf{Example Prompt WH-Question} \rule{0pt}{3.6ex}\rule[-3.2ex]{0pt}{0pt} \\ \hline

Initial Context and Questioning &
The system starts by presenting a scenario context and posing a wh-question to stimulate the learner’s thinking. & ``\textit{What aspect of the context do you find most challenging to understand?}'' \\ \hline

Response Evaluation and Feedback &
The system evaluates responses and provides hints and feedback to guide learners toward correct understanding without directly giving answers. & ``\textit{How does this part of the context relate to the overall scenario?}'' \\ \hline

Iterative Prompting &
Through iterative prompts, the system deepens the learner’s reasoning, encouraging detailed exploration and articulation. & ``\textit{Can you explain why this particular detail is significant in the scenario?}'' \\ \hline

Feedback and Exploration &
Feedback highlights correct elements and offers hints for further exploration. & ``\textit{What other factors might influence this outcome?}'' \\ \hline

Maintaining Engagement &
This approach maintains engagement through continuous, thought-provoking questions that connect new concepts to prior knowledge. & ``\textit{How would you connect this concept to what you have learned previously?}'' \\ \hline

Fostering Critical Thinking &
The system prompts learners to evaluate and critique their own responses, fostering critical thinking skills. & ``\textit{What could be a potential limitation of your current understanding?}''\\ \hline

Encouraging Reflection &
It encourages learners to reflect on their learning process and outcomes. & ``\textit{How has your understanding changed after considering this question?}'' \\ \hline
Providing Incremental Hints &
The system offers incremental hints that build upon each other to guide the learner progressively towards deeper understanding. & ``\textit{What is a simpler way to think about this problem before tackling the more complex aspects?}'' \\ \hline

Adaptive Feedback &
The feedback adapts to the learner's responses, becoming more specific as the learner's understanding develops. & ``\textit{Given your explanation, what would be the next logical step to explore?}'' \\ \hline

Encouraging Synthesis &
It encourages learners to synthesize information from different parts of the scenario to form a comprehensive understanding. & ``\textit{How can you combine these different pieces of information to solve the problem?}'' \\ \hline 

\end{tabular}
\label{tab:SocraticDialogue}
\end{table*}

The SPL features a carefully crafted prompt architecture, incorporating both the standard for constructing learning scenarios and the interactive Socratic dialogue for fostering engaging and interactive tutoring. The system dynamically refines guidance based on user input and feedback, ensuring responses are aligned with the learner's needs and learning status. This exemplifies the innovative application of dialogue-based ITSs in education. 

\subsection{System Highlights}
This system aims to provide learners with personalized, adaptive and flexible learning experiences. The main features of SPL include: 
\begin{itemize}
    \item \textbf{Personalization:} SPL creates personalized learning paths for learners, allowing them to explore different learning domains based on their interests. For example, a learner interested in psychology can choose specific topics like cognitive behavioral therapy or developmental psychology. The system traces the learner's responses and provides adaptive feedback with tailored prompt wh-questions.  
    \item \textbf{Socratic Teaching:} The system employs the Socratic teaching method, encouraging learners to think critically, reflect, and explore concepts deeply by asking thought-provoking questions instead of directly providing answers. For instance, instead of explaining the principles of cognitive dissonance directly, SPL might ask, ``What do you think happens when someone's actions contradict their beliefs?''. 
    \item \textbf{Interactivity:} SPL offers a dynamic and engaging learning experience through interactive dialogues, emulating the interactions that occur with human tutors. An example is a dialogue where the system asks, "How would you apply the concept of reinforcement in a classroom setting?" and provides feedback based on the learner's response. 
    \item \textbf{Context-Sensitivity:} SPL generates rich problem scenarios (e.g., understanding the mechanisms of attention in psychology, understanding developmental milestones in early childhood education, understanding the architecture and functioning of computer processors in computer engineering) around key concepts or knowledge components (e.g., cognitive processes, developmental stages, computer architecture) and provides guidance and feedback based on the learners' responses. 
    \item \textbf{Adaptability:} It adjusts tutoring strategies and content in response to the learner’s progress and understanding, ensuring that the learning experience is continuously optimized. For instance, if a learner struggles with a particular psychology problem, SPL might provide additional questions and multi-turn dialogues to engage the learner further. 
    \item \textbf{Cross-Domain Coverage:} The system supports learning across various domains, overcoming the limitations of many ITS that are restricted to specific fields. Examples include providing tailored content for subjects as diverse as computer science, business, engineering, psychology, nursing, mathematics, physics, and economics, etc. 
\end{itemize} 

These features foster advanced critical thinking, dynamic interactive learning, collaborative questioning, personalized learning journeys, comprehensive analytical skills, reflective metacognition, cross-disciplinary integration, and engaging motivational strategies. 

\section{System Evaluation Through Pilot Testing} 

To evaluate the SPL system's effectiveness in enhancing learner engagement, understanding, and satisfaction, we use one pilot study using the example task on essay writing. 

\subsection{Experimental Design}

This pilot testing experiment involved 10 graduate-level participants recruited from the campus. Upon entering the laboratory, participants filled out demographic information and then engaged with the SPL system for dialogue-based communication on the topic of essay writing.  Learners described their essay writing needs based on their field of study, for example: ``\textit{I am John, my major is Psychology, and I want to learn how to write empirical research papers. Please help me!}''. The 10 survey questions focused on various aspects, including the effectiveness and fluency of dialogue (\textbf{Q1}), perception of human-like interaction (\textbf{Q2}), user enjoyment (\textbf{Q3}), attractiveness of learning methods (\textbf{Q4}), happiness with learning (\textbf{Q5}),understanding enhancement (\textbf{Q6}), learning motivation (\textbf{Q7}), improvement in learning outcomes (\textbf{Q8}), satisfaction of learning needs (\textbf{Q9}), willingness to recommend the system (\textbf{Q10}), along with two open-ended questions to gather feedback. Responses were collected using a 7-point Likert scale, ranging from 1 (strongly disagree) to 7 (strongly agree). For more detailed survey questions, please refer to \hyperref[appendix:A]{Appendix A}.

The distribution of survey scores, along with their frequency percentages, was analyzed to assess various dimensions of user experience. Open-ended feedback was semantically annotated using ChatGPT (GPT-4) \cite{gilardi2023chatgpt}, with a network-based visualization highlighting similar semantic themes (using the NetworkX python package) \cite{hagberg2020networkx}. ChatGPT facilitated the precise annotation of each feedback entry, capturing the essence and complexity of responses, and identifying shared themes across different answers. This comprehensive approach allowed for a nuanced understanding of participant feedback, enhancing the evaluation of the SPL system’s performance and user satisfaction. 

\section{Results} 

\begin{figure}[!ht]
\centering
\includegraphics[height=1.9in, width=3in]{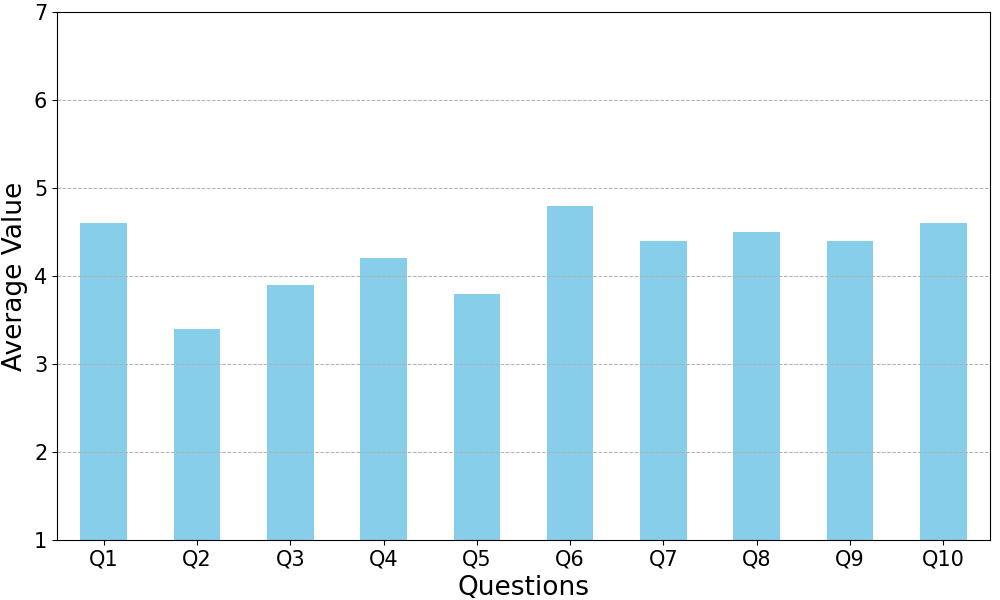}
\caption{Average Scores for \textbf{Q1} to \textbf{Q10}.} 
\label{fig:ave_distribution}
\end{figure}

Figure \ref{fig:ave_distribution} shows the average scores for \textbf{Q1} to \textbf{Q10} from the 7-point Likert scale survey conducted during the pilot experiment. The results indicate positive responses for most questions, with average scores greater than 4 for \textbf{Q1}, \textbf{Q4}, \textbf{Q6}, \textbf{Q7}, \textbf{Q8}, \textbf{Q9}, and \textbf{Q10}. These high scores suggest that participants found the system effective and engaging, particularly in terms of effectiveness and fluency of dialogue, attractiveness of learning methods, and enhancement of understanding. Additionally, the questions related to learning motivation, improvement in learning outcomes, satisfaction of learning needs, and willingness to recommend the system also received positive feedback, indicating overall satisfaction with the system's performance in these areas. Conversely, the relatively lower scores for \textbf{Q2}, \textbf{Q3}, and \textbf{Q5}, which are below 4, highlight potential areas for improvement. These questions pertain to the perception of human-like interaction, user enjoyment, and happiness with learning. The lower scores in these areas suggest that while the system is effective in delivering content and enhancing understanding, there may be a need to enhance the interactive and enjoyable aspects of the system to better engage users and make the learning experience more pleasurable. For a detailed breakdown of the scores from \textbf{Q1} to \textbf{Q10}, please refer to Figure \ref{fig:distribution} in the \hyperref[appendix:B]{Appendix B}. 

\begin{figure}[!ht]
\centering
\includegraphics[height=3in, width=3.1in]{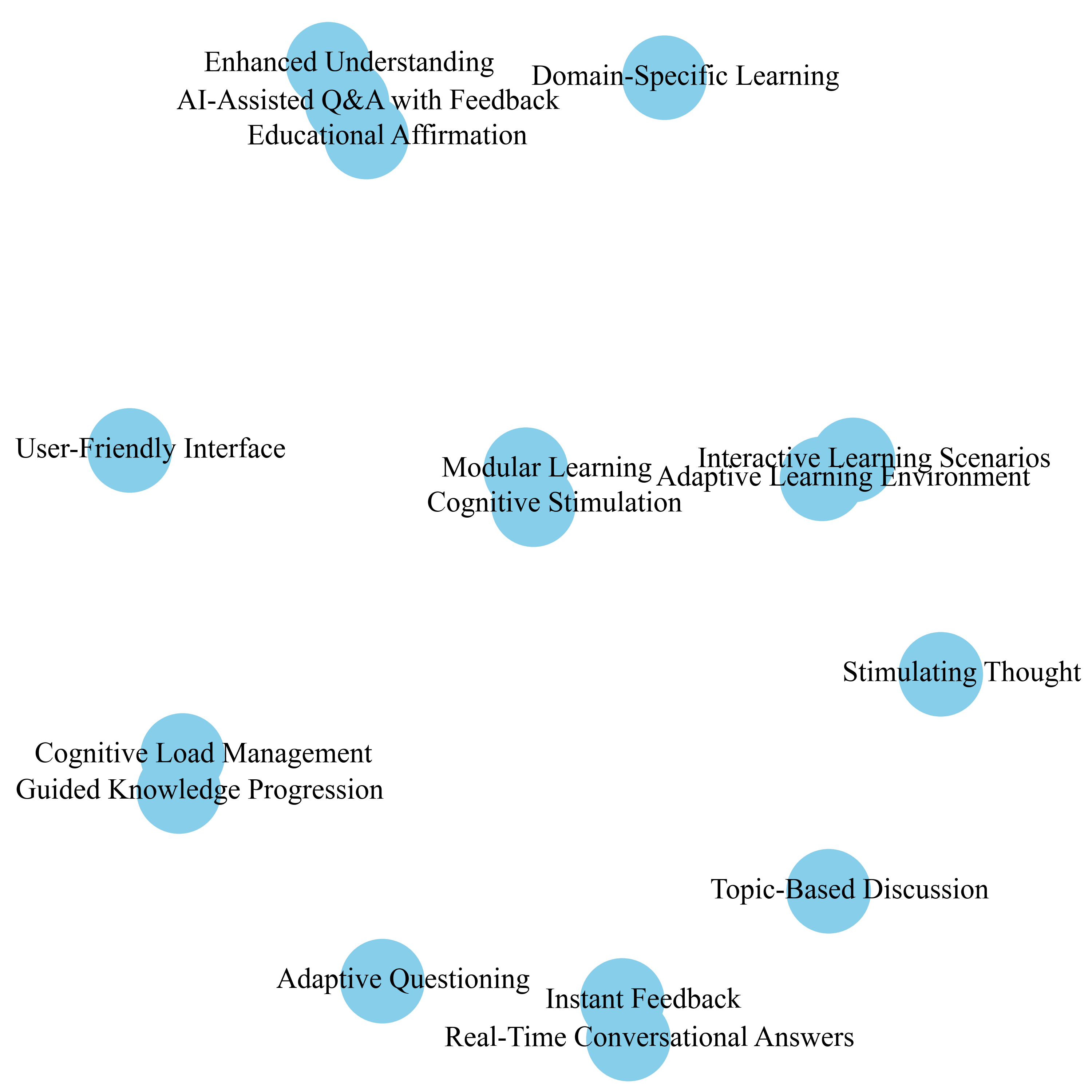}
\caption{The Semantics Visualization for Q11 about Favourite Features of SPL.}
\label{fig:open_one}
\end{figure}

Figure \ref{fig:open_one} presents the semantic annotation results and network visualization for the open question \textbf{Q11}, which investigates users' favorite features of the SPL system. Results reveal that most participants expressed positive sentiments about the system's features. Notably,  the ``AI-Assisted Q\&A with Feedback'', ``Enhanced Understanding'', and ``Educational Affirmation'' were highlighted as key advantages, underscoring the favored aspects of AI integration in learning. The other feedback also showcases the system's various strengths and benefits, with ``Interactive Learning Scenarios'' and ``Adaptive Learning Environment'' emphasize the system's flexibility and engagement, and ``Guided Knowledge Progression'' and ``Cognitive Load Management'' highlight its efficiency in organizing the learning process. Additionally, ``Real-Time Conversational Answers'' and ``Instant Feedback'' are highlighted for enhancing the interactive experience. The ``User-Friendly Interface'' and ``Topic-Based Discussion'' further improve the learning environment's usability. Overall, the feedback highlights the system's ability to deliver adaptive learning experiences, with particular appreciation for its AI-driven features and user-friendly design. 

As for the collected feedback on question \textbf{Q12}, participants emphasized the need for enhanced system performance, improved guidance, and greater clarity in the user interface. Specific issues identified included slow response times and high latency. Recommendations for improvement included the incorporation of explanatory videos and more intuitive navigation tips. Users also stressed the importance of clearer icons, such as labeled buttons, to enhance usability. They suggested that AI responses should more effectively align with previous interactions and the system should be more accessible to beginners by clearly presenting essential instructional guidance.

\section{Discussion}

Our preliminary evaluation of the SPL system has yielded promising results, demonstrating positive aspects of learner engagement and leaning experience. Participants particularly valued the system's effective use of AI to facilitate learning through the Socratic method, which could promote critical thinking and deeper comprehension. 

The system engages learners through interactive conversational process that deepens understanding, corrects misconceptions, and guides them towards their learning goals. This process is well-aligned with the expectation-misconception tailoring (EMT) principles \cite{graesser2004autotutor,graesser2005computerized,ahmed2023chatgpt}, which are designed to address and rectify learners' misconceptions effectively. By persistently guiding users with targeted prompts and feedback, the SPL system has the potential to reinforce learners' knowledge, enhance problem-solving skills, and boost their confidence. 

The SPL system enhances its educational interactions by leveraging GPT-4's capabilities. The system employs standard prompts for leasson creation and Socratic dialogue. The prompt for lesson creation organize educational scenarios by starting with broad topics and narrowing them down into specific details. This method utilizes GPT-4's strengths in reasoning, prediction, and generation to transition from general ideas to detailed learning scenarios, creating comprehensive and coherent learning experiences. Meanwhile, the prompt for interactive Socratic dialogue are carefully crafted to facilitate engaging tutoring sessions. These prompts are dynamically updated based on the flow of dialogue, ensuring that the system's responses are tailored to the learner’s current level of understanding and learning needs. Thus, the SPL system has the potential to deliver personalized and adaptive learning experiences while ensuring that the educational content is contextually appropriate.

\subsection{Limitations}

The SPL system faces several limitations that impact its overall performance and user experience. A primary concern is the time latency associated with the ChatGPT API, which can hinder the responsiveness of the system. Additionally, the implementation of learning pathways guided by the EMT approach is still limited, affecting the system's ability to fully support learners in achieving their learning goals. Another significant challenge is the hallucination issue, where the system may produce responses that are seemingly plausible but incorrect or nonsensical. Moreover, enhancing domain-specific teaching through retrieval-augmented generation remains an area for further development. Achieving truly human-like dialogue also remains difficult, with ongoing issues related to the smoothness of conversational turn-taking and latency in real-time feedback \cite{mitsui2023towards,adiwardana2020towards}. Further refinement of GPT-4 based prompt templates is needed to better assess learner states, capture responses, and compile specific knowledge source. Additionally, there is a need for generative AI models to trace and predict learner performance while exploring individual differences, building on the progress made in our previous work. \cite{zhang2024predicting,zhang20243dg,zhang2023exploring}. 

\subsection{Future Works}

To enhance the SPL system, several future work directions are proposed to improve domain-specific learning design and overall functionality. Firstly, pre-training will be utilized to incorporate expertise from educational practices and professional criteria, guiding the system towards more formal and specialized teaching methodologies. We are also exploring a scalable learning model to extend SPL as a general learning framework, capable of encompassing diverse educational domains and scenarios. This also involves integrating multimedia elements to support multimodal learning through generative AI, enhancing the system's ability to deliver rich, interactive educational experiences. 

Additionally, the development of multiple roles and agents using large language models (LLMs) will be pursued to create a more dynamic and versatile dialogue-based Intelligent Tutoring System (ITS). This will enable the system to simulate various educational roles and perspectives, providing a comprehensive learning environment. 

For the evaluation of essay submissions, future work will focus on three key areas:
\begin{itemize}
    \item Robustness of Essay Evaluation: We will assess the consistency of the evaluation process by repeatedly evaluating the same essay against a standardized rubric to ensure reliability.
    \item Sensitivity of the Evaluation: By systematically altering a well-written essay, we will evaluate how sensitive the rating system is to changes in the document, ensuring it can accurately reflect variations in quality.
    \item Psychometric Analysis of Evaluation Standards: Each evaluation standard will be treated as an individual ``person'' allowing us to analyze the effectiveness and consistency of each criterion.
    \item Evaluation of Standards: We will examine how different factors, such as the nature, type, and length of documents, influence the evaluation standards.
\end{itemize}

Following these evaluations, we aim to develop recommendations and potentially introduce a ``grading wizard'' as a user-friendly product, streamlining the grading process and enhancing user experience. This comprehensive approach aims to refine the SPL system’s educational capabilities, making it more robust, sensitive, and adaptable to a wide range of learning and assessment scenarios. 

\section{Conclusion}

In this study, we introduce the SPL system, powered by large language models (GPT-4), designed to enhance dialogue-based ITS through the Socratic method. The SPL system aims to provide personalized, adaptive, and flexible learning experiences that foster self-reflection, critical thinking, and independent thinking skills in learners. Leveraging GPT-4's prompt engineering capabilities, we employ a standard prompt for lesson creation and interactive Socratic dialogue to facilitate engaging and interactive tutoring. Preliminary pilot testing demonstrates the positive impact of SPL on learners, including increased engagement, enjoyment, and learning gains. The SPL system marks a significant improvement over traditional dialogue-based ITSs like SCHOLAR and AutoTutor, which depended on human effort for lesson design and predefined rules with limited NLP capabilities for multi-turn dialogue. Although this work is still in progress, it represents a promising step towards the next generation of dialogue-based ITS, encompassing lesson design, pedagogical strategy formulation, and the assessment of learner responses and feedback generation through generative AI. 

\section{Acknowledgments} 

% \titlenote{This collaboration project is led by Dr. Xiangen Hu from the Department of Applied Social Sciences at Hong Kong Polytechnic University.}
We would like to thank Dr. Xiangen Hu from the Department of Applied Social Sciences at Hong Kong Polytechnic University for leading this collaborative project. Additionally, we express our gratitude to Arthur C. Graesser from the Institute for Intelligent Systems at the University of Memphis for providing theoretical support and valuable insights on multi-turn dialogues, which significantly contributed to this study.


%%
%% Define the bibliography file to be used
\bibliography{references}

\begin{thebibliography}{45}
\expandafter\ifx\csname natexlab\endcsname\relax\def\natexlab#1{#1}\fi
\providecommand{\url}[1]{\texttt{#1}}
\providecommand{\href}[2]{#2}
\providecommand{\path}[1]{#1}
\providecommand{\DOIprefix}{doi:}
\providecommand{\ArXivprefix}{arXiv:}
\providecommand{\URLprefix}{URL: }
\providecommand{\Pubmedprefix}{pmid:}
\providecommand{\doi}[1]{\href{http://dx.doi.org/#1}{\path{#1}}}
\providecommand{\Pubmed}[1]{\href{pmid:#1}{\path{#1}}}
\providecommand{\bibinfo}[2]{#2}
\ifx\xfnm\relax \def\xfnm[#1]{\unskip,\space#1}\fi
%Type = Article
\bibitem[{Nye et~al.(2014)Nye, Graesser, and Hu}]{nye2014autotutor}
\bibinfo{author}{B.~D. Nye}, \bibinfo{author}{A.~C. Graesser}, \bibinfo{author}{X.~Hu},
\newblock \bibinfo{title}{Autotutor and family: A review of 17 years of natural language tutoring},
\newblock \bibinfo{journal}{International Journal of Artificial Intelligence in Education} \bibinfo{volume}{24} (\bibinfo{year}{2014}) \bibinfo{pages}{427--469}.
%Type = Article
\bibitem[{Paladines and Ramirez(2020)}]{paladines2020systematic}
\bibinfo{author}{J.~Paladines}, \bibinfo{author}{J.~Ramirez},
\newblock \bibinfo{title}{A systematic literature review of intelligent tutoring systems with dialogue in natural language},
\newblock \bibinfo{journal}{IEEE Access} \bibinfo{volume}{8} (\bibinfo{year}{2020}) \bibinfo{pages}{164246--164267}.
%Type = Article
\bibitem[{Graesser and Li(2023)}]{graesser2023intelligent}
\bibinfo{author}{A.~C. Graesser}, \bibinfo{author}{H.~Li},
\newblock \bibinfo{title}{Intelligent tutoring systems and conversational agents}  (\bibinfo{year}{2023}).
%Type = Article
\bibitem[{Graesser et~al.(2001)Graesser, VanLehn, Ros{\'e}, Jordan, and Harter}]{graesser2001intelligent}
\bibinfo{author}{A.~C. Graesser}, \bibinfo{author}{K.~VanLehn}, \bibinfo{author}{C.~P. Ros{\'e}}, \bibinfo{author}{P.~W. Jordan}, \bibinfo{author}{D.~Harter},
\newblock \bibinfo{title}{Intelligent tutoring systems with conversational dialogue},
\newblock \bibinfo{journal}{AI magazine} \bibinfo{volume}{22} (\bibinfo{year}{2001}) \bibinfo{pages}{39--39}.
%Type = Article
\bibitem[{Lin et~al.(2022{\natexlab{a}})Lin, Singh, Sha, Tan, Lang, Ga{\v{s}}evi{\'c}, and Chen}]{lin2022good}
\bibinfo{author}{J.~Lin}, \bibinfo{author}{S.~Singh}, \bibinfo{author}{L.~Sha}, \bibinfo{author}{W.~Tan}, \bibinfo{author}{D.~Lang}, \bibinfo{author}{D.~Ga{\v{s}}evi{\'c}}, \bibinfo{author}{G.~Chen},
\newblock \bibinfo{title}{Is it a good move? mining effective tutoring strategies from human--human tutorial dialogues},
\newblock \bibinfo{journal}{Future Generation Computer Systems} \bibinfo{volume}{127} (\bibinfo{year}{2022}{\natexlab{a}}) \bibinfo{pages}{194--207}.
%Type = Inproceedings
\bibitem[{Lin et~al.(2022{\natexlab{b}})Lin, Rakovic, Lang, Gasevic, and Chen}]{lin2022exploring}
\bibinfo{author}{J.~Lin}, \bibinfo{author}{M.~Rakovic}, \bibinfo{author}{D.~Lang}, \bibinfo{author}{D.~Gasevic}, \bibinfo{author}{G.~Chen},
\newblock \bibinfo{title}{Exploring the politeness of instructional strategies from human-human online tutoring dialogues},
\newblock in: \bibinfo{booktitle}{LAK22: 12th International Learning Analytics and Knowledge Conference}, \bibinfo{year}{2022}{\natexlab{b}}, pp. \bibinfo{pages}{282--293}.
%Type = Article
\bibitem[{Xu et~al.(2024)Xu, Liu, Pasupat, Kazemi et~al.}]{xu2024context}
\bibinfo{author}{X.~Xu}, \bibinfo{author}{Y.~Liu}, \bibinfo{author}{P.~Pasupat}, \bibinfo{author}{M.~Kazemi}, et~al.,
\newblock \bibinfo{title}{In-context learning with retrieved demonstrations for language models: A survey},
\newblock \bibinfo{journal}{arXiv preprint arXiv:2401.11624}  (\bibinfo{year}{2024}).
%Type = Article
\bibitem[{Lin et~al.(2024)Lin, Han, Thomas, Gurung, Gupta, Aleven, and Koedinger}]{lin2024can}
\bibinfo{author}{J.~Lin}, \bibinfo{author}{Z.~Han}, \bibinfo{author}{D.~R. Thomas}, \bibinfo{author}{A.~Gurung}, \bibinfo{author}{S.~Gupta}, \bibinfo{author}{V.~Aleven}, \bibinfo{author}{K.~R. Koedinger},
\newblock \bibinfo{title}{How can i get it right? using gpt to rephrase incorrect trainee responses},
\newblock \bibinfo{journal}{arXiv preprint arXiv:2405.00970}  (\bibinfo{year}{2024}).
%Type = Article
\bibitem[{Stamper et~al.(2024)Stamper, Xiao, and Hou}]{stamper2024enhancing}
\bibinfo{author}{J.~Stamper}, \bibinfo{author}{R.~Xiao}, \bibinfo{author}{X.~Hou},
\newblock \bibinfo{title}{Enhancing llm-based feedback: Insights from intelligent tutoring systems and the learning sciences},
\newblock \bibinfo{journal}{arXiv preprint arXiv:2405.04645}  (\bibinfo{year}{2024}).
%Type = Article
\bibitem[{Achiam et~al.(2023)Achiam, Adler, Agarwal, Ahmad, Akkaya, Aleman, Almeida, Altenschmidt, Altman, Anadkat et~al.}]{achiam2023gpt}
\bibinfo{author}{J.~Achiam}, \bibinfo{author}{S.~Adler}, \bibinfo{author}{S.~Agarwal}, \bibinfo{author}{L.~Ahmad}, \bibinfo{author}{I.~Akkaya}, \bibinfo{author}{F.~L. Aleman}, \bibinfo{author}{D.~Almeida}, \bibinfo{author}{J.~Altenschmidt}, \bibinfo{author}{S.~Altman}, \bibinfo{author}{S.~Anadkat}, et~al.,
\newblock \bibinfo{title}{Gpt-4 technical report},
\newblock \bibinfo{journal}{arXiv preprint arXiv:2303.08774}  (\bibinfo{year}{2023}).
%Type = Article
\bibitem[{Brown et~al.(2020)Brown, Mann, Ryder, Subbiah, Kaplan, Dhariwal, Neelakantan, Shyam, Sastry, Askell et~al.}]{brown2020language}
\bibinfo{author}{T.~Brown}, \bibinfo{author}{B.~Mann}, \bibinfo{author}{N.~Ryder}, \bibinfo{author}{M.~Subbiah}, \bibinfo{author}{J.~D. Kaplan}, \bibinfo{author}{P.~Dhariwal}, \bibinfo{author}{A.~Neelakantan}, \bibinfo{author}{P.~Shyam}, \bibinfo{author}{G.~Sastry}, \bibinfo{author}{A.~Askell}, et~al.,
\newblock \bibinfo{title}{Language models are few-shot learners},
\newblock \bibinfo{journal}{Advances in neural information processing systems} \bibinfo{volume}{33} (\bibinfo{year}{2020}) \bibinfo{pages}{1877--1901}.
%Type = Article
\bibitem[{Wei et~al.(2022)Wei, Wang, Schuurmans, Bosma, Xia, Chi, Le, Zhou et~al.}]{wei2022chain}
\bibinfo{author}{J.~Wei}, \bibinfo{author}{X.~Wang}, \bibinfo{author}{D.~Schuurmans}, \bibinfo{author}{M.~Bosma}, \bibinfo{author}{F.~Xia}, \bibinfo{author}{E.~Chi}, \bibinfo{author}{Q.~V. Le}, \bibinfo{author}{D.~Zhou}, et~al.,
\newblock \bibinfo{title}{Chain-of-thought prompting elicits reasoning in large language models},
\newblock \bibinfo{journal}{Advances in neural information processing systems} \bibinfo{volume}{35} (\bibinfo{year}{2022}) \bibinfo{pages}{24824--24837}.
%Type = Inproceedings
\bibitem[{Dai et~al.(2023)Dai, Lin, Jin, Li, Tsai, Ga{\v{s}}evi{\'c}, and Chen}]{dai2023can}
\bibinfo{author}{W.~Dai}, \bibinfo{author}{J.~Lin}, \bibinfo{author}{H.~Jin}, \bibinfo{author}{T.~Li}, \bibinfo{author}{Y.-S. Tsai}, \bibinfo{author}{D.~Ga{\v{s}}evi{\'c}}, \bibinfo{author}{G.~Chen},
\newblock \bibinfo{title}{Can large language models provide feedback to students? a case study on chatgpt},
\newblock in: \bibinfo{booktitle}{2023 IEEE International Conference on Advanced Learning Technologies (ICALT)}, \bibinfo{organization}{IEEE}, \bibinfo{year}{2023}, pp. \bibinfo{pages}{323--325}.
%Type = Article
\bibitem[{Kumar et~al.(2023)Kumar, Musabirov, Reza, Shi, Kuzminykh, Williams, and Liut}]{kumar2023impact}
\bibinfo{author}{H.~Kumar}, \bibinfo{author}{I.~Musabirov}, \bibinfo{author}{M.~Reza}, \bibinfo{author}{J.~Shi}, \bibinfo{author}{A.~Kuzminykh}, \bibinfo{author}{J.~J. Williams}, \bibinfo{author}{M.~Liut},
\newblock \bibinfo{title}{Impact of guidance and interaction strategies for llm use on learner performance and perception},
\newblock \bibinfo{journal}{arXiv preprint arXiv:2310.13712}  (\bibinfo{year}{2023}).
%Type = Article
\bibitem[{Park et~al.(2024)Park, Kim, Lee, Kwon, and Kim}]{park2024empowering}
\bibinfo{author}{M.~Park}, \bibinfo{author}{S.~Kim}, \bibinfo{author}{S.~Lee}, \bibinfo{author}{S.~Kwon}, \bibinfo{author}{K.~Kim},
\newblock \bibinfo{title}{Empowering personalized learning through a conversation-based tutoring system with student modeling},
\newblock \bibinfo{journal}{arXiv preprint arXiv:2403.14071}  (\bibinfo{year}{2024}).
%Type = Inproceedings
\bibitem[{Nye et~al.(2023)Nye, Mee, and Core}]{nye2023generative}
\bibinfo{author}{B.~D. Nye}, \bibinfo{author}{D.~Mee}, \bibinfo{author}{M.~G. Core},
\newblock \bibinfo{title}{Generative large language models for dialog-based tutoring: An early consideration of opportunities and concerns},
\newblock \bibinfo{year}{2023}.
%Type = Techreport
\bibitem[{Yang and Narasimhan(2023)}]{yang2023socratic}
\bibinfo{author}{R.~Yang}, \bibinfo{author}{K.~Narasimhan}, \bibinfo{title}{The socratic method for self-discovery in large language models}, \bibinfo{type}{Technical Report}, tech. rep., Princeton NLP, \bibinfo{year}{2023}.
%Type = Article
\bibitem[{Kong et~al.(2023)Kong, Yaxin, Wan, Jiang, and Wang}]{kong2023platolm}
\bibinfo{author}{C.~Kong}, \bibinfo{author}{F.~Yaxin}, \bibinfo{author}{X.~Wan}, \bibinfo{author}{F.~Jiang}, \bibinfo{author}{B.~Wang},
\newblock \bibinfo{title}{Platolm: Teaching llms via a socratic questioning user simulator}  (\bibinfo{year}{2023}).
%Type = Inproceedings
\bibitem[{Feng et~al.(2021)Feng, Magana, and Kao}]{feng2021systematic}
\bibinfo{author}{S.~Feng}, \bibinfo{author}{A.~J. Magana}, \bibinfo{author}{D.~Kao},
\newblock \bibinfo{title}{A systematic review of literature on the effectiveness of intelligent tutoring systems in stem},
\newblock in: \bibinfo{booktitle}{2021 IEEE frontiers in education conference (fie)}, \bibinfo{organization}{IEEE}, \bibinfo{year}{2021}, pp. \bibinfo{pages}{1--9}.
%Type = Incollection
\bibitem[{D'Mello and Graesser(2023)}]{d2023intelligent}
\bibinfo{author}{S.~K. D'Mello}, \bibinfo{author}{A.~Graesser},
\newblock \bibinfo{title}{Intelligent tutoring systems: How computers achieve learning gains that rival human tutors},
\newblock in: \bibinfo{booktitle}{Handbook of educational psychology}, \bibinfo{publisher}{Routledge}, \bibinfo{year}{2023}, pp. \bibinfo{pages}{603--629}.
%Type = Article
\bibitem[{Carbonell(1970)}]{carbonell1970ai}
\bibinfo{author}{J.~R. Carbonell},
\newblock \bibinfo{title}{Ai in cai: An artificial-intelligence approach to computer-assisted instruction},
\newblock \bibinfo{journal}{IEEE transactions on man-machine systems} \bibinfo{volume}{11} (\bibinfo{year}{1970}) \bibinfo{pages}{190--202}.
%Type = Article
\bibitem[{Graesser et~al.(1999)Graesser, Wiemer-Hastings, Wiemer-Hastings, Kreuz, Group et~al.}]{graesser1999autotutor}
\bibinfo{author}{A.~C. Graesser}, \bibinfo{author}{K.~Wiemer-Hastings}, \bibinfo{author}{P.~Wiemer-Hastings}, \bibinfo{author}{R.~Kreuz}, \bibinfo{author}{T.~R. Group}, et~al.,
\newblock \bibinfo{title}{Autotutor: A simulation of a human tutor},
\newblock \bibinfo{journal}{Cognitive Systems Research} \bibinfo{volume}{1} (\bibinfo{year}{1999}) \bibinfo{pages}{35--51}.
%Type = Inproceedings
\bibitem[{VanLehn et~al.(2002)VanLehn, Jordan, Ros{\'e}, Bhembe, B{\"o}ttner, Gaydos, Makatchev, Pappuswamy, Ringenberg, Roque et~al.}]{vanlehn2002architecture}
\bibinfo{author}{K.~VanLehn}, \bibinfo{author}{P.~W. Jordan}, \bibinfo{author}{C.~P. Ros{\'e}}, \bibinfo{author}{D.~Bhembe}, \bibinfo{author}{M.~B{\"o}ttner}, \bibinfo{author}{A.~Gaydos}, \bibinfo{author}{M.~Makatchev}, \bibinfo{author}{U.~Pappuswamy}, \bibinfo{author}{M.~Ringenberg}, \bibinfo{author}{A.~Roque}, et~al.,
\newblock \bibinfo{title}{The architecture of why2-atlas: A coach for qualitative physics essay writing},
\newblock in: \bibinfo{booktitle}{Intelligent Tutoring Systems: 6th International Conference, ITS 2002 Biarritz, France and San Sebastian, Spain, June 2--7, 2002 Proceedings 6}, \bibinfo{organization}{Springer}, \bibinfo{year}{2002}, pp. \bibinfo{pages}{158--167}.
%Type = Inproceedings
\bibitem[{Aleven et~al.(2004)Aleven, Ogan, Popescu, Torrey, and Koedinger}]{aleven2004evaluating}
\bibinfo{author}{V.~Aleven}, \bibinfo{author}{A.~Ogan}, \bibinfo{author}{O.~Popescu}, \bibinfo{author}{C.~Torrey}, \bibinfo{author}{K.~Koedinger},
\newblock \bibinfo{title}{Evaluating the effectiveness of a tutorial dialogue system for self-explanation},
\newblock in: \bibinfo{booktitle}{Intelligent Tutoring Systems: 7th International Conference, ITS 2004, Macei{\'o}, Alagoas, Brazil, August 30-September 3, 2004. Proceedings 7}, \bibinfo{organization}{Springer}, \bibinfo{year}{2004}, pp. \bibinfo{pages}{443--454}.
%Type = Inproceedings
\bibitem[{Rus et~al.(2014)Rus, Stefanescu, Niraula, and Graesser}]{rus2014deeptutor}
\bibinfo{author}{V.~Rus}, \bibinfo{author}{D.~Stefanescu}, \bibinfo{author}{N.~Niraula}, \bibinfo{author}{A.~C. Graesser},
\newblock \bibinfo{title}{Deeptutor: towards macro-and micro-adaptive conversational intelligent tutoring at scale},
\newblock in: \bibinfo{booktitle}{Proceedings of the first ACM conference on Learning@ scale conference}, \bibinfo{year}{2014}, pp. \bibinfo{pages}{209--210}.
%Type = Inproceedings
\bibitem[{Rus et~al.(2015)Rus, Niraula, and Banjade}]{rus2015deeptutor}
\bibinfo{author}{V.~Rus}, \bibinfo{author}{N.~Niraula}, \bibinfo{author}{R.~Banjade},
\newblock \bibinfo{title}{Deeptutor: An effective, online intelligent tutoring system that promotes deep learning},
\newblock in: \bibinfo{booktitle}{Proceedings of the AAAI Conference on Artificial Intelligence}, volume~\bibinfo{volume}{29}, \bibinfo{year}{2015}.
%Type = Inproceedings
\bibitem[{Ahmed et~al.(2023)Ahmed, Shubeck, and Hu}]{ahmed2023chatgpt}
\bibinfo{author}{F.~Ahmed}, \bibinfo{author}{K.~Shubeck}, \bibinfo{author}{X.~Hu},
\newblock \bibinfo{title}{Chatgpt in the generalized intelligent framework for tutoring},
\newblock in: \bibinfo{booktitle}{Proceedings of the 11th Annual Generalized Intelligent Framework for Tutoring (GIFT) Users Symposium (GIFTSym11)}, \bibinfo{organization}{US Army Combat Capabilities Development Command--Soldier Center}, \bibinfo{year}{2023}, p. \bibinfo{pages}{109}.
%Type = Article
\bibitem[{Schmucker et~al.(2023)Schmucker, Xia, Azaria, and Mitchell}]{schmucker2023ruffle}
\bibinfo{author}{R.~Schmucker}, \bibinfo{author}{M.~Xia}, \bibinfo{author}{A.~Azaria}, \bibinfo{author}{T.~Mitchell},
\newblock \bibinfo{title}{Ruffle\&riley: Towards the automated induction of conversational tutoring systems},
\newblock \bibinfo{journal}{arXiv preprint arXiv:2310.01420}  (\bibinfo{year}{2023}).
%Type = Article
\bibitem[{Schmucker et~al.(2024)Schmucker, Xia, Azaria, and Mitchell}]{schmucker2024ruffle}
\bibinfo{author}{R.~Schmucker}, \bibinfo{author}{M.~Xia}, \bibinfo{author}{A.~Azaria}, \bibinfo{author}{T.~Mitchell},
\newblock \bibinfo{title}{Ruffle\&riley: Insights from designing and evaluating a large language model-based conversational tutoring system},
\newblock \bibinfo{journal}{arXiv preprint arXiv:2404.17460}  (\bibinfo{year}{2024}).
%Type = Article
\bibitem[{Abu-Rasheed et~al.(2024)Abu-Rasheed, Abdulsalam, Weber, and Fathi}]{abu2024supporting}
\bibinfo{author}{H.~Abu-Rasheed}, \bibinfo{author}{M.~H. Abdulsalam}, \bibinfo{author}{C.~Weber}, \bibinfo{author}{M.~Fathi},
\newblock \bibinfo{title}{Supporting student decisions on learning recommendations: An llm-based chatbot with knowledge graph contextualization for conversational explainability and mentoring},
\newblock \bibinfo{journal}{arXiv preprint arXiv:2401.08517}  (\bibinfo{year}{2024}).
%Type = Misc
\bibitem[{Dan et~al.(2023)Dan, Lei, Gu, Li, Yin, Lin, Ye, Tie, Zhou, Wang, Zhou, Zhou, Chen, Zhou, He, and Qiu}]{dan2023educhat}
\bibinfo{author}{Y.~Dan}, \bibinfo{author}{Z.~Lei}, \bibinfo{author}{Y.~Gu}, \bibinfo{author}{Y.~Li}, \bibinfo{author}{J.~Yin}, \bibinfo{author}{J.~Lin}, \bibinfo{author}{L.~Ye}, \bibinfo{author}{Z.~Tie}, \bibinfo{author}{Y.~Zhou}, \bibinfo{author}{Y.~Wang}, \bibinfo{author}{A.~Zhou}, \bibinfo{author}{Z.~Zhou}, \bibinfo{author}{Q.~Chen}, \bibinfo{author}{J.~Zhou}, \bibinfo{author}{L.~He}, \bibinfo{author}{X.~Qiu}, \bibinfo{title}{Educhat: A large-scale language model-based chatbot system for intelligent education}, \bibinfo{year}{2023}. \href{http://arxiv.org/abs/2308.02773}{{\tt arXiv:2308.02773}}.
%Type = Article
\bibitem[{Dai et~al.(????)Dai, Tsai, Lin, Aldino, Jin, Li, Gasevic et~al.}]{daiassessing}
\bibinfo{author}{W.~Dai}, \bibinfo{author}{Y.-S. Tsai}, \bibinfo{author}{J.~Lin}, \bibinfo{author}{A.~Aldino}, \bibinfo{author}{F.~Jin}, \bibinfo{author}{T.~Li}, \bibinfo{author}{D.~Gasevic}, et~al.,
\newblock \bibinfo{title}{Assessing the proficiency of large language models in automatic feedback generation: An evaluation study}  (????).
%Type = Misc
\bibitem[{Lin et~al.(2024)Lin, Han, Thomas, Gurung, Gupta, Aleven, and Koedinger}]{lin2024i}
\bibinfo{author}{J.~Lin}, \bibinfo{author}{Z.~Han}, \bibinfo{author}{D.~R. Thomas}, \bibinfo{author}{A.~Gurung}, \bibinfo{author}{S.~Gupta}, \bibinfo{author}{V.~Aleven}, \bibinfo{author}{K.~R. Koedinger}, \bibinfo{title}{How can i get it right? using gpt to rephrase incorrect trainee responses}, \bibinfo{year}{2024}. \href{http://arxiv.org/abs/2405.00970}{{\tt arXiv:2405.00970}}.
%Type = Article
\bibitem[{Zhang et~al.(2024)Zhang, Lin, Borchers, Sabatini, Hollander, Cao, and Hu}]{zhang2024predicting}
\bibinfo{author}{L.~Zhang}, \bibinfo{author}{J.~Lin}, \bibinfo{author}{C.~Borchers}, \bibinfo{author}{J.~Sabatini}, \bibinfo{author}{J.~Hollander}, \bibinfo{author}{M.~Cao}, \bibinfo{author}{X.~Hu},
\newblock \bibinfo{title}{Predicting learning performance with large language models: A study in adult literacy},
\newblock \bibinfo{journal}{arXiv preprint arXiv:2403.14668}  (\bibinfo{year}{2024}).
%Type = Misc
\bibitem[{{Xiangen Hu}(2024)}]{SPL_FAQ}
\bibinfo{author}{{Xiangen Hu}}, \bibinfo{title}{{FAQ About SPL}}, \bibinfo{howpublished}{\url{https://spl.skoonline.org/FAQ/index.html?lang=en}}, \bibinfo{year}{2024}.
%Type = Article
\bibitem[{Koshik(2003)}]{koshik2003wh}
\bibinfo{author}{I.~Koshik},
\newblock \bibinfo{title}{Wh-questions used as challenges},
\newblock \bibinfo{journal}{Discourse Studies} \bibinfo{volume}{5} (\bibinfo{year}{2003}) \bibinfo{pages}{51--77}.
%Type = Article
\bibitem[{Fox and Thompson(2010)}]{fox2010responses}
\bibinfo{author}{B.~A. Fox}, \bibinfo{author}{S.~A. Thompson},
\newblock \bibinfo{title}{Responses to wh-questions in english conversation},
\newblock \bibinfo{journal}{Research on Language and Social Interaction} \bibinfo{volume}{43} (\bibinfo{year}{2010}) \bibinfo{pages}{133--156}.
%Type = Article
\bibitem[{Gilardi et~al.(2023)Gilardi, Alizadeh, and Kubli}]{gilardi2023chatgpt}
\bibinfo{author}{F.~Gilardi}, \bibinfo{author}{M.~Alizadeh}, \bibinfo{author}{M.~Kubli},
\newblock \bibinfo{title}{Chatgpt outperforms crowd workers for text-annotation tasks},
\newblock \bibinfo{journal}{Proceedings of the National Academy of Sciences} \bibinfo{volume}{120} (\bibinfo{year}{2023}) \bibinfo{pages}{e2305016120}.
%Type = Article
\bibitem[{Hagberg and Conway(2020)}]{hagberg2020networkx}
\bibinfo{author}{A.~Hagberg}, \bibinfo{author}{D.~Conway},
\newblock \bibinfo{title}{Networkx: Network analysis with python},
\newblock \bibinfo{journal}{URL: https://networkx. github. io}  (\bibinfo{year}{2020}).
%Type = Article
\bibitem[{Graesser et~al.(2004)Graesser, Lu, Jackson, Mitchell, Ventura, Olney, and Louwerse}]{graesser2004autotutor}
\bibinfo{author}{A.~C. Graesser}, \bibinfo{author}{S.~Lu}, \bibinfo{author}{G.~T. Jackson}, \bibinfo{author}{H.~H. Mitchell}, \bibinfo{author}{M.~Ventura}, \bibinfo{author}{A.~Olney}, \bibinfo{author}{M.~M. Louwerse},
\newblock \bibinfo{title}{Autotutor: A tutor with dialogue in natural language},
\newblock \bibinfo{journal}{Behavior Research Methods, Instruments, \& Computers} \bibinfo{volume}{36} (\bibinfo{year}{2004}) \bibinfo{pages}{180--192}.
%Type = Article
\bibitem[{Graesser et~al.(2005)Graesser, Hu, and McNamara}]{graesser2005computerized}
\bibinfo{author}{A.~C. Graesser}, \bibinfo{author}{X.~Hu}, \bibinfo{author}{D.~S. McNamara},
\newblock \bibinfo{title}{Computerized learning environments that incorporate research in discourse psychology, cognitive science, and computational linguistics.}  (\bibinfo{year}{2005}).
%Type = Article
\bibitem[{Mitsui et~al.(2023)Mitsui, Hono, and Sawada}]{mitsui2023towards}
\bibinfo{author}{K.~Mitsui}, \bibinfo{author}{Y.~Hono}, \bibinfo{author}{K.~Sawada},
\newblock \bibinfo{title}{Towards human-like spoken dialogue generation between ai agents from written dialogue},
\newblock \bibinfo{journal}{arXiv preprint arXiv:2310.01088}  (\bibinfo{year}{2023}).
%Type = Article
\bibitem[{Adiwardana et~al.(2020)Adiwardana, Luong, So, Hall, Fiedel, Thoppilan, Yang, Kulshreshtha, Nemade, Lu et~al.}]{adiwardana2020towards}
\bibinfo{author}{D.~Adiwardana}, \bibinfo{author}{M.-T. Luong}, \bibinfo{author}{D.~R. So}, \bibinfo{author}{J.~Hall}, \bibinfo{author}{N.~Fiedel}, \bibinfo{author}{R.~Thoppilan}, \bibinfo{author}{Z.~Yang}, \bibinfo{author}{A.~Kulshreshtha}, \bibinfo{author}{G.~Nemade}, \bibinfo{author}{Y.~Lu}, et~al.,
\newblock \bibinfo{title}{Towards a human-like open-domain chatbot},
\newblock \bibinfo{journal}{arXiv preprint arXiv:2001.09977}  (\bibinfo{year}{2020}).
%Type = Article
\bibitem[{Zhang et~al.(2024)Zhang, Lin, Borchers, Cao, and Hu}]{zhang20243dg}
\bibinfo{author}{L.~Zhang}, \bibinfo{author}{J.~Lin}, \bibinfo{author}{C.~Borchers}, \bibinfo{author}{M.~Cao}, \bibinfo{author}{X.~Hu},
\newblock \bibinfo{title}{3dg: A framework for using generative ai for handling sparse learner performance data from intelligent tutoring systems},
\newblock \bibinfo{journal}{arXiv preprint arXiv:2402.01746}  (\bibinfo{year}{2024}).
%Type = Inproceedings
\bibitem[{Zhang et~al.(2023)Zhang, Pavlik~Jr, Hu, Cockroft, Wang, and Shi}]{zhang2023exploring}
\bibinfo{author}{L.~Zhang}, \bibinfo{author}{P.~I. Pavlik~Jr}, \bibinfo{author}{X.~Hu}, \bibinfo{author}{J.~L. Cockroft}, \bibinfo{author}{L.~Wang}, \bibinfo{author}{G.~Shi},
\newblock \bibinfo{title}{Exploring the individual differences in multidimensional evolution of knowledge states of learners},
\newblock in: \bibinfo{booktitle}{International Conference on Human-Computer Interaction}, \bibinfo{organization}{Springer}, \bibinfo{year}{2023}, pp. \bibinfo{pages}{265--284}.

\end{thebibliography}

\appendix

%Appendix A
\section{Survey Table} \label{appendix:A}
\begin{table}[h!]
\centering
\caption{Survey Questions for SPL System Evaluation}
\footnotesize
\renewcommand{\arraystretch}{1.3}
\begin{tabular}{p{1.5cm}|p{5.4cm}}
\hline
\textbf{Question No.} & \textbf{Survey Question} \\ \hline

Q1 & I believe the dialogue in the SPL is effective and smooth (1 = strongly disagree, 7 = strongly agree; same below). \\ \hline
Q2 & I feel like I am interacting with a person in the SPL. \\ \hline
Q3 & I enjoy learning in the SPL. \\ \hline
Q4 & I find the learning methods provided by the SPL attractive. \\ \hline
Q5 & I feel happy while learning in the SPL. \\ \hline
Q6 & The SPL helps me understand the learning content. \\ \hline
Q7 & I am motivated to learn in the SPL. \\ \hline
Q8 & Learning in the SPL can improve my current knowledge performance. \\ \hline
Q9 & I feel that the SPL meets my learning needs. \\ \hline
Q10 & I am willing to recommend the SPL to others. \\ \hline
Q11 & What is your favorite feature or function of the system? \\ \hline
Q12 & What other feedback or suggestions do you have for the system? \\ \hline

\end{tabular}
\label{tab:SurveyQuestions}
\end{table}

As shown in Table \ref{tab:SurveyQuestions}
, all the survey questions following the 7-point Likert scale are presented. These questions were designed to evaluate various aspects of the SPL system, including its effectiveness, user interaction, enjoyment, and overall satisfaction. By using a 7-point Likert scale, we aimed to capture a wide range of participant responses, from strong disagreement to strong agreement, providing a nuanced understanding of their experiences. Additionally, two open-ended questions were included to gather qualitative feedback, allowing participants to elaborate on their favorite features and provide suggestions for improvement. This comprehensive approach ensures a thorough evaluation of the SPL system from multiple perspectives. 

\section{Percentage Distribution of Scores for Q1 to Q10} \label{appendix:B}

\begin{figure}[!ht]
\centering
\includegraphics[height=3in, width=3in]{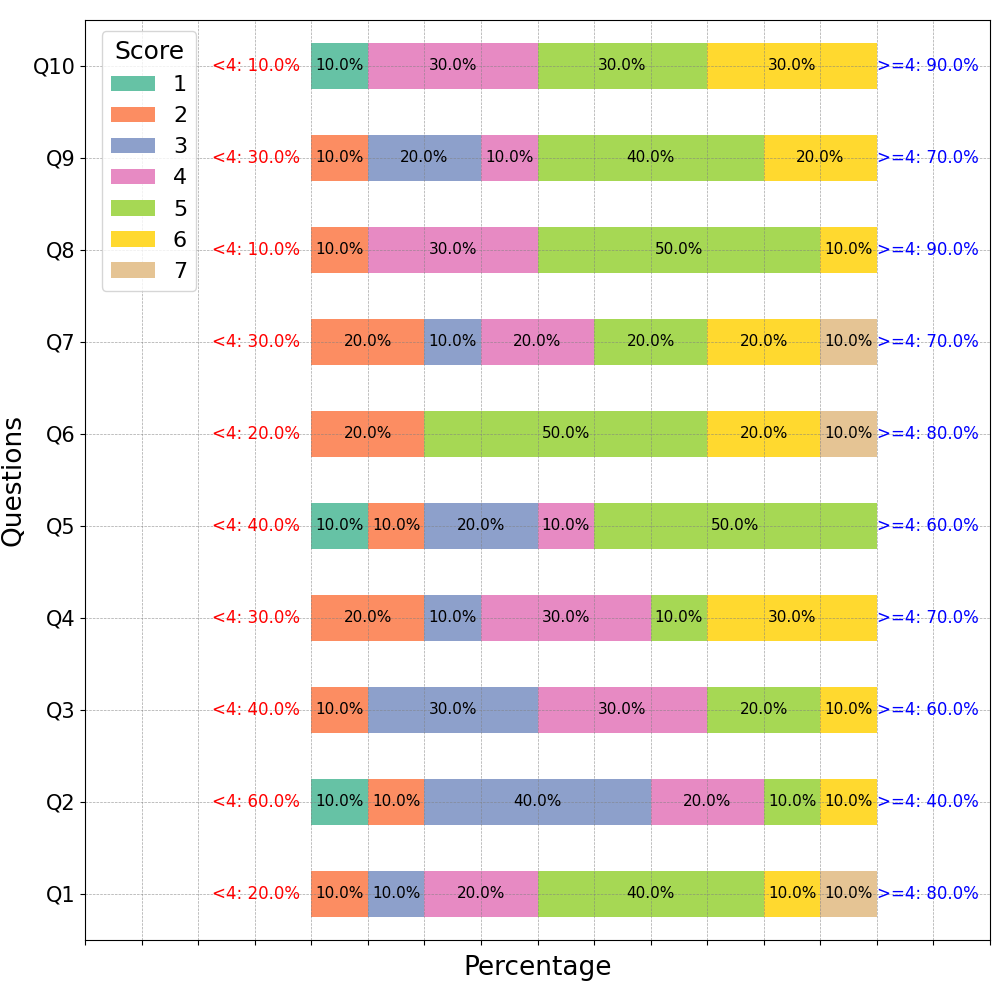}
\caption{Percentage Distribution of Scores for Q1 to Q10.} 
\label{fig:distribution}
\end{figure}

Figure \ref{fig:distribution} shows the percentage distribution of scores for \textbf{Q1} to \textbf{Q10}, centered around a score of 4, the neutral point on the 7-point Likert scale. Generally, most participants indicated positive impacts of the SPL system, as the total percentage of scores above 4 exceeds those below 4, as shown by the labeled percentages on the left and right sides of the bars for each question. The results suggest that the system's performance is above average, with approximately 28\% of scores below 4 and 72\% of scores 4 or above. Additionally, the average score for each question is predominantly above 4, with the exception of \textbf{Q2}, which assesses the human-likeness of the SPL system, indicating it is perceived as less human-like. Examining the individual questions, \textbf{Q8} (improvement in learning outcomes) and \textbf{Q10} ( willingness to recommend the system) have the highest percentage (both are 90\%) of scores in the 5-7 range, suggesting strong positive feedback in these areas. \textbf{Q6} (understanding enhancement) also shows 80\% of responses in the positive range (50\% at score 5, 20\% at score 6, and 10\% at score 7). On the other hand, \textbf{Q2} stands out with a significant portion of responses in the lower range (40\% at score 3, 10\% at score 2, and 10\% at score 1), suggesting a critical area for improvement regarding the perceived human-likeness of the system. 

\end{document}